\documentclass[10pt, conference]{IEEEtran}
\IEEEoverridecommandlockouts

\usepackage{url}

\ifCLASSINFOpdf
  \usepackage[pdftex]{graphicx}
\else
  \usepackage[dvips]{graphicx}
\fi
\usepackage{fixltx2e}
\usepackage{url}

\begin{document}
%
\title{A Smart System for Selection of  Optimal Product Images in E-Commerce}


\author{\IEEEauthorblockN{Abon Chaudhuri, Paolo Messina, Samrat Kokkula, Aditya Subramanian,\\
Abhinandan Krishnan, Shreyansh Gandhi, Alessandro Magnani, Venkatesh Kandaswamy}
\IEEEauthorblockA{Walmart Labs, Sunnyvale, CA, USA\\
achaudhuri,venky@walmartlabs.com\\
}

}


%


\maketitle

\begin{abstract}
%
In e-commerce, content quality of the product catalog plays a key role in delivering a satisfactory experience to the customers. In particular, visual content such as product images influences customers' engagement and purchase decisions. With the rapid growth of e-commerce and the advent of artificial intelligence, traditional content management systems are giving way to automated scalable systems. In this paper, we present a machine learning driven visual content management system for extremely large e-commerce catalogs. For a given product, the system aggregates images from various suppliers, understands and analyzes them to produce a superior image set with optimal image count and quality, and arranges them in an order tailored to the demands of the customers. The system makes use of an array of technologies, ranging from deep learning to traditional computer vision, at different stages of analysis. In this paper, we outline how the system works and discuss the unique challenges related to applying machine learning techniques to real-world data from e-commerce domain. We emphasize how we tune state-of-the-art image classification techniques to develop solutions custom made for a massive, diverse, and constantly evolving product catalog. We also provide the details of how we measure the system's impact on various customer engagement metrics.
\end{abstract}


%
\IEEEpeerreviewmaketitle

\section{Introduction}
\label{sec:intro}
Images play a key role in influencing the quality of customer experience and the customers' decision-making path in e-commerce transactions. Images provide detailed product information that helps the customer build confidence in the product quality and fulfillment promises. Additionally, images provide inspirational cues about the experience associated with the product e.g. lifestyle images that show how the customer can enjoy the product. In most cases, the impact of images in the customers' journey from product discovery to evaluation and decision-making is greater than the impact of other types of product content such as description and product attributes (brand, size etc.). 

\begin{figure}[ht]
    \centering
     \includegraphics[width=\linewidth]{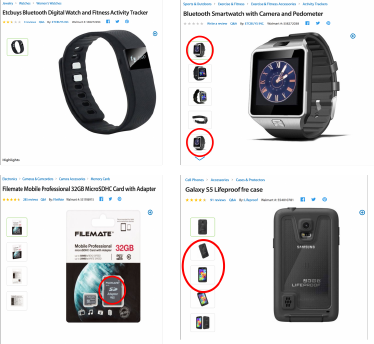}
     \caption{Examples of common problems with product images. \textbf{Top Left:} The product has just one image. That is too few for the customer for making any decision. \textbf{Top Right:} The watch has too many images where the first one and the fifth one (circled in red) are duplicates \textbf{Bottom Left:} Every image shows the product inside packaging. A clear picture of the memory card (circled in red) is not provided to the customer. \textbf{Bottom Right:} This is an example of incorrectly ordered images. The back of the cell phone is presented to the customer before the front or the screen view.}
     \label{fig:problems}
\end{figure}

While a nicely curated image set can significantly elevate the customer experience, bad or incorrect images or an incoherent set of images can severely hinder the customers' progress towards making decisions. It can, in fact, break the emotional connection between a customer and a product. Figure~\ref{fig:problems} showcases some of the problems commonly found with product images. Too few of them, too many of them, irrelevant images, inappropriate images, duplicates, incorrectly ordered images are some of the prominent problems. Hence, it is very important that the product images satisfy the product images:
\begin{itemize}
\item Each individual image passes quality and compliance standards (examples of low-quality images are blurry or distorted ones or images with misleading or distracting patterns and text. Images with offensive or violent content are considered non-compliant.)
\item Each individual image offers useful information about the product (for example, a picture of the entire living room is not so useful when the customer is looking to buy a small end table. A picture of a car is not needed if the customer is looking for a car interior polish)
\item Each image is significantly different from the rest (for example, multiple pictures of a TV taken from almost the same angle cause cognitive overload on the customer)
\item The images reveal the information to customers in meaningful steps e.g. in an order that is easy to process (for example, showing the back of a TV before showing the front leads to poor customer experience)
\end{itemize}

The traditional technique of manual curation of images and other content by in-house experts or crowd workers does not scale for product catalogs containing millions of items. Human errors in compiling product information and limitations of software systems severely hinder the ability to provide a homogeneous content experience across categories to the customer. Hence, it is important to employ the human workforce to generate training data and develop machine learning  based techniques using that data to ensure that each of the above-mentioned criteria is satisfied for all product images. In this paper, we provide a concise account of such a system that uses computer vision and deep learning to create the optimal set of images for products. 

This paper offers a solution to the above mentioned business problem. The models and algorithms presented in the paper are based on known machine learning techniques. However, the paper discusses innovative customizations needed to deal with very large datasets, noisy and sparse labels, and ad-hoc business requirements.
%
%
\section{Smart Image Selection System}
\label{sec:framework}
\begin{figure*}[tb]
    \centering
     \includegraphics[width=0.9\linewidth]{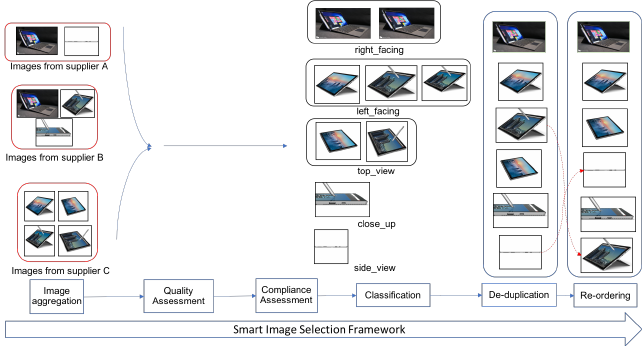}
     \caption{\textbf{Smart Image Selection Framework} explained with an example. The framework first assembles 9 images for a tablet from 3 suppliers, namely A, B and C. Each of them is first sent through quality and compliance checks. The images that survive the filters are then classified into 5 groups. One or two images from each groups is selected for the final set. Finally, the images in the final set are re-ordered. For example, the additional right facing images (3$^{rd}$ from top) is pushed down the list while the side view (at the very bottom) is promoted up. The goal is to produce a final image set which is better than each input set of images.}
     \label{fig:framework}
\end{figure*}
We present a smart image selection system that creates an optimal image set that provides adequate and relevant information in an orderly fashion for every product in the catalog. The system works through the following stages:
\begin{enumerate}
\item \emph{Image Aggregation:} In a very large catalog, many items are supplied and sold by multiple parties. Each one of them may provide a number of images (see leftmost column of Figure~\ref{fig:framework} for an example), but none of them may have an exhaustive set of images. The image aggregation module simply collects all the images from various content providing parties and product suppliers together. The number of images collected this way is often quite higher than the number of images supplied by an individual supplier.
\item \emph{Image Quality and Compliance Assessment:} This module contains deep learning classifiers that examine each image in the aggregated set for quality issues (such as blurriness, low resolution, poor use of real estate and compliance issues (such as adult and violent content, fake marketing badges). The images that fails such standards are discarded.
\item \emph{Image Classification:} In this stage, the images are classified into different types based on viewpoints and some other criteria. To give an example, the common types of images for a tee shirt would be a picture of the front, a picture of the back, a lifestyle images of a human model wearing it, and a close-up of the texture. For some product categories, the accessories are an expected type of image (for example, picture of a stylus that comes with a tablet). The output of this stage is a set of images grouped by types.
\item \emph{Image De-duplication:} The de-duplication module selects from each group a set of representative images (usually just one). More than one image can be chosen from a group if it contains images with subtle yet important differences. Figure~\ref{fig:framework} shows an example where two images from \textit{left\_facing} views are chosen since one of the contains the stylus. This module makes sure that all the chosen images are sufficiently different from one another.
\item \emph{Image Re-ordering} Finally, the selected images are re-ordered to gradually unfold the product information to the customer, moving from getting the big picture to looking into important details. 
\end{enumerate}

In the following sections, we explain the details of each technical module of the smart image selection framework.

\subsection{Image Quality Assessment}
\label{subsec:quality}
A number of image quality issues are known to have a negative impact on the customers~\cite{qualitypatent}. Some of the issues are explicit such as low-resolution images that appear blurry on large screen. Some quality issues are more intrinsic such as images where the central object is surrounded by excessive white or colored space. 

The smart image system detects a number of image quality problems using a combination of deep learning classifiers. At present, it consists of three classifiers: the first one detects blurry images, the second one detects images with small objects relative to their background area and the third one detects images with temporary placeholder material with no real content 

Each problem is addressed with a binary classifier that is fine-tuned with product image data. However, the main challenge in building these classifiers lies in the sparsity of the training data. Certain types of low-quality product images appear so infrequently that it is very difficult to create a sizable training dataset without manually scanning through millions of images. 

We solve the data sparsity problem by employing clever data augmentation techniques to synthetically generate problematic images. A few example augmentations are: 
\begin{itemize} 
\item Regular images are cropped keeping the central object and then resized back to the original size to create a dataset of blurry images
\item Regular images are padded with whitespace of different amounts in different directions to create their low-quality counterparts
\item Traditional computer vision techniques such as graph cut segmentation~\cite{grabcut} is used to locate the central object in an image. Then the background is replaced by disproportionately high amount of white space or natural background.
\end{itemize}

With the training dataset at hand, we fine-tune the last few layers of Inception V3 networks pre-trained on our own catalog data. We keep augmenting the dataset iteratively until the desired precision and recall is achieved.
\begin{figure}[ht]
    \centering
     \includegraphics[width=\linewidth]{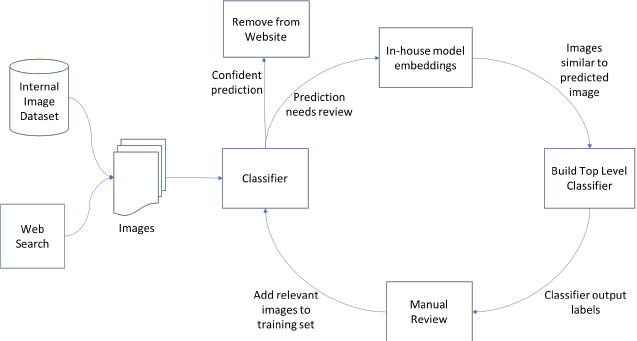}
     \caption{\textbf{Iterative Training Framework:} It allows to  start from a very small amount of training data and gradually improve the model by adding examples found by the model back to the training data.}
     \label{fig:retrain_classifier}
\end{figure}

\subsection{Non-compliant Image Detection and Removal}
\label{subsec:offensive}
Non-complaint images cover a broad spectrum of images that contain adult, violent, racially inappropriate content. Clearly, such images can severely damage the customer experience, lead to legal issues, and invite complaints from advertising platforms. All of these eventually cause erosion of the company brand. 

We approach the detection of such images as a supervised image classification problem. However, the main challenge lies in the sparsity of training examples. It is extremely hard to build a training dataset with a handful of examples for each type of non-compliant problems.

We solve this problem by combining three approaches: collecting examples from the internet, creating synthetic examples whenever possible, and most importantly, iterative augmentation of the training data. As Figure~\ref{fig:retrain_classifier} shows, the first set of training images are collected from internal product catalog and by web search. The first version of the classifier is built using this dataset. Then, a moderate amount of images are run through the classifier with the hope of detecting more positive examples. If the classifier finds more such examples with high confidence, they are removed from the website and added to the training data as well. If the classifier finds examples with low confidence, they are sent for manual review by crowdsourcing teams. In either case, the detected image is passed through an deep image similarity model (trained on our product images) to obtain a number of similar images. The model prediction and the obtained similar images are sent to crowdsourcing teams for manual review. The responses sent by the crowd workers are added to the training set and the classifier is retrained. A number of iterations are usually necessary to obtain desired accuracy. 

%
%
%
\subsection{Image Type Classification}
\label{subsec:tagging}
This step classifies the images of a product into a number of types that are expected for that type of product. For example, the popular types for laptop images front view, back view, top view, side view and close-up of various parts. Broadly, the types are characterized by how the product looks from different views, how its features work, and how it is used or experienced by the customers. 

This image classification serves as an important precursor to image ranking. Our system first needs to detect different views of a product from its images in order to rank them.

We treat this as a supervised classification problem because we determine the expected image types for a given product category are finite in advance. The data analysts study a fraction of product images from a category and formulate a list of image types that are commonly found for a category. We then create a small training dataset of images from that category and have them labeled via crowd-sourcing. 

\begin{figure}[ht]
    \centering
     \includegraphics[width=\linewidth]{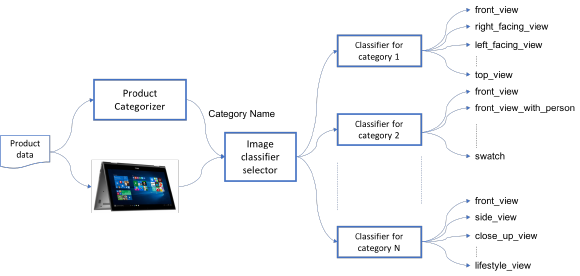}
     \caption{\textbf{Image Classification Module:} A deep learning classifier is trained for each product category.}
     \label{fig:image_tagger}
\end{figure}

As shown in Figure~\ref{fig:image_tagger}, we train an image classifier for each category (such as Television, Tablet Computer, Laptop etc.). Since the catalog contains thousands of categories and each category has a number of types, developing a single model to address all categories would be a daunting task. Instead, we leverage the in-house product categorizer that predicts the category of the item from its title and description. Our system then directs items to different classifiers based on the product's category.

\begin{figure}[ht]
    \centering
     \includegraphics[width=\linewidth]{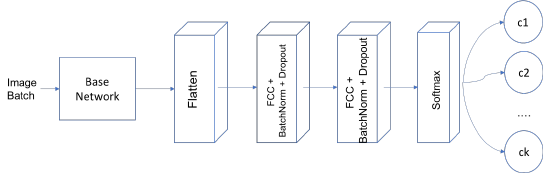}
     \caption{\textbf{Architecture of Image Classification Module:} A shallow network is added on top of popular deep learning architectures for each product category.}
     \label{fig:image_model}
\end{figure}

We have a relatively small dataset of images for training the model for each category. While a category may have as many as a million items in it, the training dataset usually contains few thousands images at best. This is why we leverage open source deep neural networks pre-trained on Imagenet dataset~\cite{imagenet_cvpr09}. We have experimented with two approaches: 
\begin{itemize}
\item One approach is to re-train the last few layers of a pre-trained network with our images 
\item Another approach is to add a few fully connected layers on top of a pre-trained network (Figure~\ref{fig:image_model}) and to train the additional layers. In our experiment, we train to minimize categorical cross-entropy using stochastic gradient descent with a slow learning rate as the optimizer. 
\end{itemize}
Probably because we use small datasets to retrain each classifier, the second approach has consistently showed marginally better performance. 

\subsection{Image De-duplication}
\label{subsec:dedup}
\begin{figure}[ht]
    \centering
     \includegraphics[width=\linewidth]{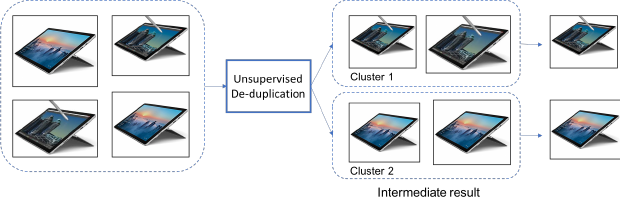}
     \caption{\textbf{Unsupervised De-duplication algorithm:} Images with the same class label are run through an unsupervised clustering to create a de-duplicated set of images.}
     \label{fig:dedup_flow}
\end{figure}

The image classification step (Section~\ref{subsec:tagging}) produces a number of groups of images. Each group usually contains several images of the same type that are exactly identical or nearly identical (example: 4 left facing images in Figure~\ref{fig:dedup_flow}). We run each group of images through an unsupervised clustering algorithm that performs pairwise comparisons and creates one or more clusters within the group. The images within a cluster are almost identical, they vary either by size, resolution, or the white space around the central object. Since they contain the same content, selecting any one from each cluster is sufficient for our objective of optimizing the customer experience. On the other hand, the images in different clusters are of the same type, but their contents are significantly different (example, the left view and the left view with a stylus). Hence, selecting one image from each cluster leads to a de-duplicated set.
\begin{figure}[ht]
    \centering
     \includegraphics[width=0.85\linewidth]{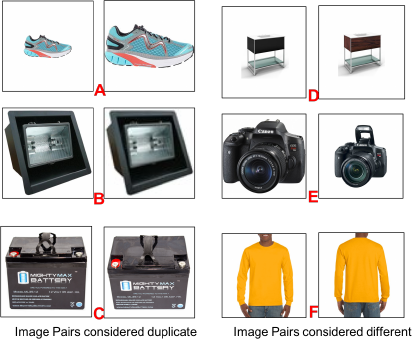}
     \caption{\textbf{Examples of identical and different pairs:} Pairs A, B, and C represent real duplicate pairs from the product catalog. D, E and F represents very similar pairs that should be considered as different images and both should be presented to the customer.}
     \label{fig:dedup_examples}
\end{figure}

\textbf{Details of Clustering:} Given an image group, we do not know the number of clusters in advance, hence we cannot use off-the-shelf algorithms such as K-means. Instead, we create the clusters in a top-down fashion by comparing image pairs. A pairwise comparator that accepts two images and returns a binary decision - duplicate (covers both exact and near duplicates) or different - is at the heart of our clustering algorithm. Identifying \textit{near-duplicate} images is a reasonably difficult problem in computer vision. It is more difficult in our case because the definition of near duplicate varies widely from category to category. In some cases, two images that are almost identical can represent two different features of the same product and hence, should be considered different. Figure~\ref{fig:dedup_examples} shows such pairs (D, E, F) on the right column. For example, a picture of a camera and another picture from the same angle with the flash popped out are very similar-looking images, but both of them should be retained for the customer to view. On the other hand, the examples shown on the left column (A, B, C) are also near duplicate images, but they should be treated as duplicates since they present the exact same information to the customer. Only one from such pairs should be shown to the customer.

\textbf{Details of Pairwise Comparator:} Since our image comparator needs to address such subtle differences, a supervised method would require us to collect training data for every category, if not every product. Considering the difficulty and the cost, we consider the option of using an unsupervised technique that would work without the knowledge of what is in the image. We observe that in most cases where two images should be treated as different, there are sub-regions with strong differences between the edges (such as E and F) or there is a significant difference in color (example pair D). Based on this observation, we choose to develop custom descriptors that are compared using some distance metric. Using embeddings generated by deep neural nets as descriptors is another option, but such embeddings are tuned for object recognition and classification, hence they tend to be oblivious to subtle differences between two images of the same object.
\begin{figure}[ht]
    \centering
     \includegraphics[width=\linewidth]{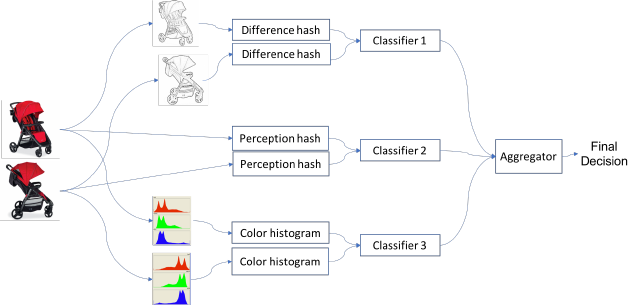}
     \caption{\textbf{Image pair comparator:} A combination of hash and histogram based techniques is used to detect near-duplicate images}
     \label{fig:dedup_algo}
\end{figure}

We leverage image hashes that are well known as image descriptors or signatures (usually 64 bits long). In short, a hash is a signature of an image after reduction of size and color. We leverage the fact that certain variations of image hashes are more robust to subtle edge differences. We employ a combination of two hash-based descriptors and a histogram based descriptor (Figure~\ref{fig:dedup_algo}):
\begin{itemize}
\item \textbf{Perceptual Hash based Comparator:} Perceptual hash~\cite{phash, phash2} of an image is computed by performing cosine transform on a color and size reduced version of the image. Two images that are almost identical should have almost equal perceptual hashes. This comparator uses the difference between the perceptual hashes and applies a threshold on them to return a binary decision based on distance.
\item \textbf{Difference Hash based Comparator:} Difference hash works similarly, but it emphasizes the differences in the edges. This is effective in catching differences shown in cases E and F in Figure~\ref{fig:dedup_examples}. We use edge-enhanced versions of the product images as input to accentuate the difference further.
\item \textbf{Histogram based Comparator:} Since hash-based descriptors are applied after color reduction, they fail to differentiate between two images of the same product in two different colors (as shown in the example D in Figure~\ref{fig:dedup_examples}). This descriptor is designed to catch those cases. It uses HSV color histograms of the images as the descriptor and computes Chi-square distance between the histograms. A threshold is used to turn the distance to a binary decision.
\end{itemize}
Two images are considered duplicate if the decisions from all three classifiers indicate are same. The rationale behind this is as follows: from the customer's point of view, the cost of wrongly dropping an image is higher than leaving a near duplicate. Hence, our de-duplicator is optimized for minimizing false positives.
\subsection{Image Re-ordering}
\label{subsec:ranking}
The de-duplicated set is a list of images along with image types. We re-order this list based on category specialists' (in-house product experts) experience and other A/B tests performed in-house. These tests give us a prioritized list of image types for a category. If more than one image of a type is present, the additional ones are pushed down the list. Ideally, images can be re-ordered based on customer clicks on images, if that data is available. 
\section{Results}
\label{sec:results}
In this section we present quantitative results from some of the key algorithmic components followed by qualitative case studies. 
\subsection{Quantitative Results}
\label{sec:quant}
While we experiment using various models and architectures, the techniques that are in use can broadly be divided into two categories:
\begin{itemize} 
\item Techniques where a shallow neural network or a different model such as logistic regression is trained on top of embeddings taken from the last dense layer of a general-purpose deep learning model
\item Techniques where the last few layers of a pre-trained deep learning model is re-trained with our data
\end{itemize} 
In our experience, the quality and quantity of the data makes a much stronger difference than the choice of the modeling technique or the deep learning architecture. All of the following experiments are run on internal datasets, and the goal of the experiments is not to beat the state-of-the-art, rather to find the method that best suits our data. Hence, the absolute performance numbers are not presented, instead in each table we compare a number of candidate techniques against the best performing one. For each table, the reference method (usually the best performing one) is shown in bold.

\textbf{Non-compliant Image Detection Performance:} The results provided here are from a model that separates images containing assault rifles from the rest. This is a  specific use case of non-compliant image detection (Section~\ref{subsec:offensive}
). In the \textbf{first approach}, an Inception-v3 based deep learning model is trained on a very large dataset of our catalog images for a different problem - product categorization. Then, the embeddings generated from this model are passed through either a logistic regression or a random forest model to perform the assault rifle detection. The same embeddings are re-used for various classification tasks. In the \textbf{second approach}, we use a Resnet50 model and retrain its last convolution block. We also experiment by retraining the last inception module of a pre-trained Inception-v3 model. We perform the retraining with $\sim$150 positive examples (assault rifles) and $\sim$1M negative examples.
\begin{table}[h!]
\centering
\caption{Non-compliant Image Detection Results}
\begin{tabular}{|p{2.2cm}|p{1.6cm} c c c|} 
\hline
Method/Network & Pre-trained on & Precision & Recall & F1-Score \\ 
\hline\hline
Deep embeddings + logistic regression & our catalog images & -0.02 & -0.11  & -0.06\\ 
\hline
Deep embeddings + random forest & our catalog images & -0.05 & -0.11  & -0.08\\
\hline
Resnet-50 & Imagenet & +0.01 & -0.07  & -0.03\\
\hline
\textbf{Inception-V3} & Imagenet & x & x  & x\\ 
\hline
\end{tabular}
\label{table:weapon_results}
\end{table}

Results from both approaches are presented in Table~\ref{table:weapon_results}. Precision and recall values of the second approach are marginally better. Advantage of first approach is that the entire model is trained on our own images for a general task, hence it may generalize well for new data and new tasks. However, retraining this model is a heavyweight task. The second approach allows quick retraining with small data, but the model is primarily trained on Imagenet dataset.

\textbf{Image Type Classifier Performance:} Table~\ref{table:image_type2} presents the performances of the image type classifier (described in Section~\ref{subsec:tagging} for two categories based on a 10\% hold-out set. The first approach of building a shallow network on top of pre-trained architectures shows slightly better performance over retraining the architecture itself, possibly because we have very small datasets (1000 to 3000 images per category) for retraining.
\begin{table}[htb]
\centering
\caption{Image Classification Performance for \textit{Tablet Computers}}
\begin{tabular}{|p{3.4cm}|c c c|} 
\hline
Technique & Precision & Recall & F1-Score\\ 
\hline
\hline
\textbf{VGG19 + shallow network} & x & x & x\\ 
Resnet50 + shallow network & -0.01 & -0.0 & -0.0\\ 
Inception + shallow network & -0.05 & -0.04 & -0.04\\
\hline
Resnet50 (retrained 1 block) & +0.0 & +0.01 & +0.0\\
Inception (retrained 1 block) & -0.06 & -0.05 & -0.05\\
\hline
\end{tabular}

\caption{Image Classification Performance for for \textit{T-Shirts}}
\begin{tabular}{|p{3.4cm}|c c c|} 
\hline
Technique & Precision & Recall & F1-Score\\ 
\hline
\hline
\textbf{VGG19 + shallow network} & x & x & x\\ 
Resnet50 + shallow network & +0.0 & +0.0 & +0.0\\ 
Inception + shallow network & -0.03 & -0.03 & -0.03\\
\hline
Resnet50 (retrained 1 block) & +0.0 & +0.0 & +0.0\\
Inception (retrained 1 block) & -0.01 & -0.01 & -0.01\\
\hline
\end{tabular}
\label{table:image_type2}
\end{table}

\begin{figure}[ht]
    \centering
     \includegraphics[width=0.75\linewidth]{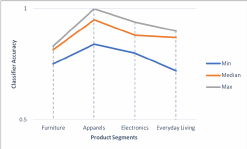}
     \caption{\textbf{Average Performance of classifiers in different segments (verticals) of the catalog:} The min, max, and median accuracy of image type classifiers from four distinct areas of the catalog}
     \label{fig:model_by_area}
\end{figure}

Accuracy of this classifier for a category largely depends on the quality of the data. Hence, even if it is possible to develop highly accurate classifiers for a few categories, it is worthwhile to study how the performances of these classifiers hold across categories, or at a higher level, across verticals such as fashion or electronics. A vertical encompasses a number of categories falling into the same business domain. Figure~\ref{fig:model_by_area} presents the average performance of the classifiers from four major areas. For each vertical, the plot shows the min, max, and the median accuracy of the classifiers from within the vertical. If a vertical (such as furniture in this case) trails behind the rest, it indicates some potential issue with the data, or it may indicate that the categories under that vertical are not lending themselves well to the model chosen. In that case, we plan targeted model tuning for that vertical.

\textbf{Image Pair De-duplicator Performance:} We measure the performance of the image pair comparator (described in Section~\ref{subsec:dedup}) on a benchmark dataset that incorporates the standard cases and the specific examples that are more common in our dataset. The dataset consists of 4400 image pairs with labels different or duplicate. About half of the duplicate pairs are synthetically generated and the rest are real examples hand-picked from the catalog. We have randomly incorporated geometric transformations of the main object (translation, rotation, shear etc.) or color-space transformations (such as sharpness and contrast) to real images to produce their near-duplicate counterparts. 

The image pair comparator first computes the distance between two images. Then it applies a threshold to the distance to generate the final label. We have attempted this in a few different ways:
\begin{itemize}
\item Compute cosine similarity directly on the images
\item Compute cosine similarity between deep embeddings generated by standard deep learning models 
\item Convert the images into image hashes and then compute Hamming distance between the hashes 
\end{itemize}
For each method, we determine the most optimal threshold by testing a range of thresholds on the benchmark dataset. In a way, the benchmark dataset serves as an indirect way to learn the threshold.
%
%
\begin{table}[tb]
\centering
\begin{tabular}{|c p{1.2cm}| c c c |} 
\hline
Technique & Optimal Threshold & Precision & Recall & F1-Score\\ 
\hline
\hline
Cosine Similarity & 0.99 & +0.0 & -0.18 & -0.1\\
Avg. Hash & 10 & -0.07 & -0.05 & -0.01\\ 
Perception Hash & 10 & -0.07 & +0.0 & -0.04\\ 
Difference Hash & 20 & -0.09 & +0.01 & -0.03\\ 
Wavelet Hash & 15 & -0.12 & -0.04 & -0.09\\ 
\textbf{Hash Ensemble} & 20 & x & x & x\\
\hline
VGG19 + cosine & 0.85 & -0.09 & -0.15 & -0.12\\
Inception + cosine & 0.85 & -0.23 & -0.2 & -0.21\\
Resnet50 + cosine & 0.85 & -0.10 & -0.21 & -0.15\\
\hline
\end{tabular}
\caption{Performance of Pairwise De-duplication}
\label{table:dedup_results}
\end{table}

Table~\ref{table:dedup_results} presents a range of techniques starting from computing cosine similarity on the images to a number of hash based techniques to using cosine similarity on deep embeddings. It turns out that the Hash ensemble technique proposed by us addresses our business-specific use cases the best whereas deep learning based embeddings tend to ignore the fine differences and focus on the central object.

\subsection{Qualitative Results}
\label{sec:qual}
\begin{table*}[tb!]
\centering
\begin{tabular}{|c|c|c|} 
\hline
Product Category & Existing product images in order (from left to right) & Recommended product images in order (from left to right)\\ 
\hline
\hline
Laptop & \includegraphics[width=0.38\textwidth]{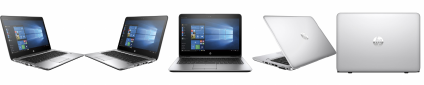} & \includegraphics[width=0.38\textwidth]{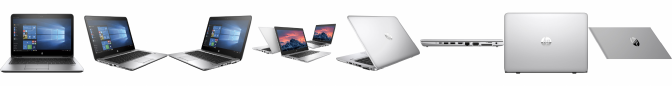}\\
\hline
Tablet & \includegraphics[width=0.38\textwidth]{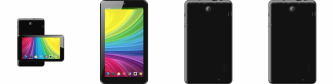} & \includegraphics[width=0.38\textwidth]{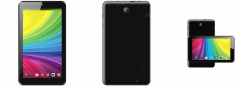}\\
\hline
Sofa & \includegraphics[width=0.38\textwidth]{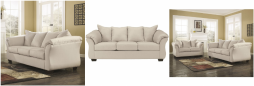} & \includegraphics[width=0.38\textwidth]{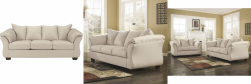}\\
\hline
\end{tabular}
\caption{Before-after scenarios from three products of different types. Second column shows the images that are currently being displayed in order (from left to right). Third column displays the images recommended by our system in order (from left to right) \textbf{Top Row:} Our system adds new images to this product. \textbf{Middle Row:} Our system removes duplicates and hence reduces the number of images. \textbf{Bottom Row:} Our system suggests a new order of the images.}
\label{table:overall_results}
\end{table*}
The final outcome of our system is an ordered list of product images. Accuracy of the individual models discussed so far are intermediate steps that help create a more meaningful final result. We present a few actual final results produced by our system in Table~\ref{table:overall_results} to enable side-to-side comparison of the current product images and the recommended ones.

\textbf{Case Study 1:} The \textbf{top row of Table~\ref{table:overall_results}} shows an example of a laptop which originally had 5 images. Our smart image system recommends a list of 8 images by aggregating 3 new images from other suppliers. One of the new images is the side view of the laptop ($6^{th}$ from left) that shows the available inlets and outlets. Also, in the e-commerce domain, the first image (also called the \textit{hero image}) conveys very high importance to the customer experience. For this product, our system identifies that a front view of the laptop is available, but it is being shown as the $3^{rd}$ image. Since it is a better candidate for the hero image compared to the side-facing views, it moves it up to make it the first image.

\textbf{Case Study 2:} The \textbf{middle row of Table~\ref{table:overall_results}} shows an example of a tablet computer for which the smart image selection system removes one image. The $3^{rd}$ and the $4^{th}$ images of the item (in the middle column) are clearly duplicates, hence our system removes one. It also identifies the current $2^{nd}$ image as a front view and a better candidate to be the hero image. Hence, it moves it to the first spot. The new order of images is front-back-multiview (from left to right). Since multiview is anyway a composite of front and back views, it is not offering a lot of new information to the customer. Hence, it is pushed down. 

\textbf{Case Study 3:} The \textbf{bottom row of Table~\ref{table:overall_results}} presents an example from a different domain - furniture. In this example, our system keeps the number of images intact, but moves the $2^{nd}$ picture, which is a clear front view of the sofa with a pure white background, to the $1^{st}$ spot. 
\subsection{Measurement of Impact on Customers} 
\label{sec:causal}
A/B testing is the right method for measuring customer impact of such a system. However, performing it at a large-scale in a production environment is often not feasible. Network effects impact the test, and a segment of the customers are actually impacted by the test. Hence, choose to estimate the impact using counterfactual analysis, a form of causal impact analysis~\cite{causalimpact}, that generates similar results without having to create a real control set. At a very high level, the method estimates the impact of a change or intervention on a set of items (in our case changing the image set is the intervention) compared to another set of items (known as a synthetic control set) with similar behavioral history (in our case items with similar search and browse history). 

We have used this technique to obtain a statistical estimate of the impact of our system on two metrics:  add-to-cart (ATC) and conversion (or actual purchase). We studies these metrics for a period of 3 weeks after the intervention. We have observed an overall positive impact of this system on most of the categories. Table~\ref{table:counterfactual} presents the results for 4 categories. The table shows the estimated average relative causal effect in the second column (performance of the intervened items compared to the control items) along with a posterior probability (the third column). 
\begin{table}[tb]
\centering
\caption{Counterfactual Analysis of Customer Metrics}
\begin{tabular}{|c p{2.5cm} p{2.3cm}|} 
\hline
Category & Relative effect on Add-to-Cart (\%) & Probability of causal effect (\%) \\ 
\hline
\hline
Laptops & -1.6 & 64\\
Tablets & 24 & 99\\
Televisions & 9 & 93\\
Monitors & 27 & 99\\
\hline
Category & Relative effect on Conversion(\%) & Probability of causal effect (\%) \\ 
\hline
Laptops & 5.6 & 79\\
Tablets & 34 & 99\\
Televisions & 20 & 99\\
Monitors & 39 & 99\\
\hline
\end{tabular}
\label{table:counterfactual}
\end{table}
As the table suggests, most estimates are positive except a mild negative impact on ATC of laptops. However, it may be noted that the posterior probability of that estimate is significantly low, suggesting that the effect may be spurious and would generally not be considered statistically significant. On the other hand, the positive effects associated with 99\% probability indicates that the probability of obtaining this effect by chance is very small and the effect can be considered statistically significant.

Another consistent observation from this analysis is that for a given category, the effect on conversion is usually stronger than the effect on ATC. Our hypothesis is that as we provide a more informative and complete image set, the customer behavior shows more certainty. If a customer adds a product to the cart, they are more likely to purchase it. On the other hand, poor or incomplete imagery often weakens the customer's confidence, leading to indecision or change of mind even after adding an item to cart. 

In another experiment run on a set of tires (Table~\ref{table:wearout}), we verify an assumption that the initial effect of improving the images wears out with time and eventually stabilizes. This highlights the importance of periodic re-evaluation of product images as new images become available.
\begin{table}[t]
\centering
\caption{Effect of Time on Causal Effect}
\begin{tabular}{|c p{2.2cm} p{2.2cm}|} 
\hline
Number of weeks & Relative effect on Add-to-Cart (\%) & Probability of causal effect (\%) \\ 
\hline
\hline
2 & 6.9 & 99\\
4 & 4.2 & 99\\
6 & 1.7 & 96\\
\hline
\end{tabular}
\label{table:wearout}
\end{table}
All the inferences from causal impact analysis depend critically on the assumption that the covariates (factors other than images for a product) were not themselves affected by the intervention. This definitely holds in our case. However, the model also assumes that the relationship between covariates and treated time series, as established during the pre-intervention period, remains stable throughout the post-intervention period. This may not hold strictly in the real world where various events may impact the time series behavior. Even after considering a margin of error in the analysis, the overall trend observed is positive in most cases.
\subsection{Performance of the Deployed System} 
\label{sec:system}
The image selection system is currently deployed as a web service to our distributed environment. The web service is hosted as a python+Apache+Flask framework. It exposes a REST end point that is accessible to various product processing pipelines. 

It may be noted that the time that our system takes to process a single item is a factor of how many images are found by the image aggregation module. The entire processing (image download + classification + de-duplication + ranking) takes 1 to several seconds depending on the number of images. The downloading of the images from URLs to the memory is parallelized. Inference on deep learning models are run in batches, whenever possible.

Even with these optimizations, processing image workloads is often not real time. However, it is acceptable if the improved images appear after a few seconds of a product's appearance on the website. Hence, the recommendations made by our system are asynchronously merged with the product data and the images get updated in a few seconds. Also, not every item in the product catalog is a candidate for this system. For example, we can safely bypass items with no or just one image, items that have already been manual curated etc. For any new category, we first perform a mock run on a small sample of items to obtain an estimate of the average number of images per item and the average percentage of items that benefit from our algorithms. Then, compute resources are allocated based on that estimate.

\section{Related Work}
\label{sec:relwork}
The importance of images in e-commerce is a well-studied problem. A study by Chen et al.~\cite{chen2013} establishes a link between images and purchase intentions. The results presented by Di et al.~\cite{image_ecom_14} show positive evidence that better images can lead to increase in buyer's attention, trust and conversion rate. A recent study by Zakrewsky et al.~\cite{qualitypopularity} shows correlation between popular items and images with high quality images. Our system addresses a number of image-related problems using concepts from image classification and similarity detection. 

Image quality assessment (IQA) is an well-established area~\cite{qualitysurvey}. Our scenario falls in the domain of with-reference quality assessment because in most cases the low quality images are well defined. 

We leverage the recent advances in deep learning research for image classification. Starting from Alexnet~\cite{alexnet}, deep neural net models have been established as the state-of-the-art for image classification and object recognition. A number of deep learning architectures such as VGG16 and VGG19~\cite{vgg16}, residual network~\cite{resnet50}, Inception v1~\cite{inceptionv1}, v2, and v3~\cite{inceptionv2} have been proposed since then. More recently, B. Zoph et al.~\cite{nasnet} have proposed Nasnet that is able to learn the best suited architecture by itself. In this paper we start with one of VGG, Resnet or Inception architectures pre-trained on Imagenet~\cite{imagenet_cvpr09} dataset, and fine tune the weights with our dataset. 

Our image de-duplication module determines if two images are near duplicates or have significant differences between them. Hashes or similar image signatures are sometimes used to rank visual search results~\cite{simpatent}. In the deep learning domain, Siamese networks~\cite{siamese} are often used for this purpose. Such networks learn latent feature representations of two images using shared model weights and compare the representations. If a positive and a negative match for each image is available, such networks can be trained to minimize triplet loss~\cite{triplet}. 

The problem of non-compliant image detection suffers from sparsity of training data. One-shot learning~\cite{oneshot} addresses this by comparing a new image against a set of known examples.    

\section{Conclusion}
\label{sec:conclusion}
In this paper we present a machine-learning driven system that delivers superior online shopping experience by selecting optimal product images. We outline the solution framework and explain the core technical steps in detail. Most importantly, we highlight the challenges of working with large and noisy datasets, describe the adjustments and tweaks made to the data and the modeling techniques to address the scale.

The presented system is already deployed in production and it has processed hundreds of thousands of products. However, as the data and customer behavior keep evolving and new technologies arrive, we constantly try to leverage them to improve our system. For example, we plan to learn the most optimal order of images from customer behavior. We plan to enhance the analysis of the system by tracking image-specific customer metrics such as number of clicks on an image, or the time spent on an image. We also plan to quantify the relevance of an image by understanding the product title and description.






\bibliographystyle{IEEEtran}
\bibliography{main}

\begin{thebibliography}{10}
\providecommand{\url}[1]{#1}
\csname url@samestyle\endcsname
\providecommand{\newblock}{\relax}
\providecommand{\bibinfo}[2]{#2}
\providecommand{\BIBentrySTDinterwordspacing}{\spaceskip=0pt\relax}
\providecommand{\BIBentryALTinterwordstretchfactor}{4}
\providecommand{\BIBentryALTinterwordspacing}{\spaceskip=\fontdimen2\font plus
\BIBentryALTinterwordstretchfactor\fontdimen3\font minus
  \fontdimen4\font\relax}
\providecommand{\BIBforeignlanguage}[2]{{%
\expandafter\ifx\csname l@#1\endcsname\relax
\typeout{** WARNING: IEEEtran.bst: No hyphenation pattern has been}%
\typeout{** loaded for the language `#1'. Using the pattern for}%
\typeout{** the default language instead.}%
\else
\language=\csname l@#1\endcsname
\fi
#2}}
\providecommand{\BIBdecl}{\relax}
\BIBdecl

\bibitem{qualitypatent}
\BIBentryALTinterwordspacing
A.~Goswami, N.~Chittar, A.~Dasdan, S.~P. Ghatare, S.~N. Gaikwad, and S.~H.
  Chung, ``Image quality assessment to merchandise an item,'' Patent
  8\,675\,957, 03 18, 2014. [Online]. Available:
  \url{https://patents.google.com/patent/US8675957B2/en}
\BIBentrySTDinterwordspacing

\bibitem{grabcut}
\BIBentryALTinterwordspacing
C.~Rother, V.~Kolmogorov, and A.~Blake, ``"grabcut": Interactive foreground
  extraction using iterated graph cuts,'' \emph{ACM Trans. Graph.}, vol.~23,
  no.~3, pp. 309--314, Aug. 2004. [Online]. Available:
  \url{http://doi.acm.org/10.1145/1015706.1015720}
\BIBentrySTDinterwordspacing

\bibitem{imagenet_cvpr09}
J.~Deng, W.~Dong, R.~Socher, L.-J. Li, K.~Li, and L.~Fei-Fei, ``{ImageNet: A
  Large-Scale Hierarchical Image Database},'' in \emph{CVPR09}, 2009.

\bibitem{phash}
``Perceptual image hashes,''
  \url{https://jenssegers.com/61/perceptual-image-hashes}, 2014.

\bibitem{phash2}
``phash - the open source perceptual hash library,''
  \url{http://www.phash.org/}, 2014.

\bibitem{causalimpact}
K.~H. Brodersen, F.~Gallusser, J.~Koehler, N.~Remy, and S.~L. Scott,
  ``Inferring causal impact using bayesian structural time-series models,''
  \emph{Annals of Applied Statistics}, vol.~9, pp. 247--274, 2015.

\bibitem{chen2013}
M.-Y. Chen and C.-I. Teng, ``A comprehensive model of the effects of online
  store image on purchase intention in an e-commerce environment,''
  \emph{Electronic Commerce Research}, vol.~13, no.~1, pp. 1--23, Mar 2013.

\bibitem{image_ecom_14}
\BIBentryALTinterwordspacing
W.~Di, N.~Sundaresan, R.~Piramuthu, and A.~Bhardwaj, ``Is a picture really
  worth a thousand words?: - on the role of images in e-commerce,'' in
  \emph{Proceedings of the 7th ACM International Conference on Web Search and
  Data Mining}, ser. WSDM '14.\hskip 1em plus 0.5em minus 0.4em\relax New York,
  NY, USA: ACM, 2014, pp. 633--642. [Online]. Available:
  \url{http://doi.acm.org/10.1145/2556195.2556226}
\BIBentrySTDinterwordspacing

\bibitem{qualitypopularity}
\BIBentryALTinterwordspacing
S.~Zakrewsky, K.~Aryafar, and A.~Shokoufandeh, ``Item popularity prediction in
  e-commerce using image quality feature vectors,'' \emph{CoRR}, vol.
  abs/1605.03663, 2016. [Online]. Available:
  \url{http://arxiv.org/abs/1605.03663}
\BIBentrySTDinterwordspacing

\bibitem{qualitysurvey}
\BIBentryALTinterwordspacing
P.~Mohammadi, A.~Ebrahimi{-}Moghadam, and S.~Shirani, ``Subjective and
  objective quality assessment of image: {A} survey,'' \emph{CoRR}, vol.
  abs/1406.7799, 2014. [Online]. Available:
  \url{http://arxiv.org/abs/1406.7799}
\BIBentrySTDinterwordspacing

\bibitem{alexnet}
A.~Krizhevsky, I.~Sutskever, and G.~E. Hinton, ``Imagenet classification with
  deep convolutional neural networks,'' in \emph{Advances in Neural Information
  Processing Systems 25}, F.~Pereira, C.~J.~C. Burges, L.~Bottou, and K.~Q.
  Weinberger, Eds.\hskip 1em plus 0.5em minus 0.4em\relax Curran Associates,
  Inc., 2012, pp. 1097--1105.

\bibitem{vgg16}
K.~Simonyan and A.~Zisserman, ``Very deep convolutional networks for
  large-scale image recognition,'' \emph{CoRR}, vol. abs/1409.1556, 2014.

\bibitem{resnet50}
\BIBentryALTinterwordspacing
K.~He, X.~Zhang, S.~Ren, and J.~Sun, ``Deep residual learning for image
  recognition,'' \emph{CoRR}, vol. abs/1512.03385, 2015. [Online]. Available:
  \url{http://arxiv.org/abs/1512.03385}
\BIBentrySTDinterwordspacing

\bibitem{inceptionv1}
\BIBentryALTinterwordspacing
C.~Szegedy, W.~Liu, Y.~Jia, P.~Sermanet, S.~E. Reed, D.~Anguelov, D.~Erhan,
  V.~Vanhoucke, and A.~Rabinovich, ``Going deeper with convolutions,''
  \emph{CoRR}, vol. abs/1409.4842, 2014. [Online]. Available:
  \url{http://arxiv.org/abs/1409.4842}
\BIBentrySTDinterwordspacing

\bibitem{inceptionv2}
\BIBentryALTinterwordspacing
C.~Szegedy, V.~Vanhoucke, S.~Ioffe, J.~Shlens, and Z.~Wojna, ``Rethinking the
  inception architecture for computer vision,'' \emph{CoRR}, vol.
  abs/1512.00567, 2015. [Online]. Available:
  \url{http://arxiv.org/abs/1512.00567}
\BIBentrySTDinterwordspacing

\bibitem{nasnet}
\BIBentryALTinterwordspacing
B.~Zoph, V.~Vasudevan, J.~Shlens, and Q.~V. Le, ``Learning transferable
  architectures for scalable image recognition,'' \emph{CoRR}, vol.
  abs/1707.07012, 2017. [Online]. Available:
  \url{http://arxiv.org/abs/1707.07012}
\BIBentrySTDinterwordspacing

\bibitem{simpatent}
\BIBentryALTinterwordspacing
A.~Gulli', A.~Savona, T.~Yang, X.~Liu, B.~Li, A.~Choksi, F.~Tanganelli, and
  L.~Carnevale, ``Similarity detection and clustering of images,'' Patent
  US7\,801\,893B2, 09 21, 2010. [Online]. Available:
  \url{https://patents.google.com/patent/US7801893}
\BIBentrySTDinterwordspacing

\bibitem{siamese}
S.~Chopra, R.~Hadsell, and Y.~LeCun, ``Learning a similarity metric
  discriminatively, with application to face verification,'' in \emph{2005 IEEE
  Computer Society Conference on Computer Vision and Pattern Recognition
  (CVPR'05)}, vol.~1, June 2005, pp. 539--546 vol. 1.

\bibitem{triplet}
\BIBentryALTinterwordspacing
F.~Schroff, D.~Kalenichenko, and J.~Philbin, ``Facenet: {A} unified embedding
  for face recognition and clustering,'' \emph{CoRR}, vol. abs/1503.03832,
  2015. [Online]. Available: \url{http://arxiv.org/abs/1503.03832}
\BIBentrySTDinterwordspacing

\bibitem{oneshot}
\BIBentryALTinterwordspacing
L.~Fei-Fei, R.~Fergus, and P.~Perona, ``One-shot learning of object
  categories,'' \emph{IEEE Trans. Pattern Anal. Mach. Intell.}, vol.~28, no.~4,
  pp. 594--611, Apr. 2006. [Online]. Available:
  \url{https://doi.org/10.1109/TPAMI.2006.79}
\BIBentrySTDinterwordspacing

\end{thebibliography}
%



\end{document}